\documentclass[11pt]{article}

\usepackage[preprint]{acl}

\usepackage{microtype}

\usepackage[english,bidi=default]{babel} %

\usepackage{graphicx}
\usepackage{longtable}
\usepackage{colortbl}

\usepackage{multirow}
\usepackage{tabularx}

\usepackage{pdfpages}
\NewCommandCopy\oriincludepdf\includepdf
\usepackage{polyglossia}
\usepackage{amsmath}
\setmainlanguage{english}
\setotherlanguage{arabic}
\RenewCommandCopy\includepdf\oriincludepdf

\usepackage{fontspec}
\newfontfamily\arabicfont[
  Script=Arabic,
  Scale=1,
  Ligatures=TeX,
  BoldFont = Amiri-Bold.ttf,
  ItalicFont = Amiri-Italic.ttf,
  BoldItalicFont = Amiri-BoldItalic.ttf
]{Amiri-Regular.ttf}
\newfontfamily\arabicfonttt[Script=Arabic,Scale=1]{Amiri-Regular.ttf}

\usepackage{array}
\usepackage{float}
\restylefloat{table}

\usepackage{booktabs}
\usepackage{euscript}
\usepackage{pifont}
\newcommand{\cmark}{\ding{51}}  %
\newcommand{\xmark}{\ding{55}}  %
\newcommand{\tmark}{$\triangle$} %

\newcommand{\para}[1]{\smallskip\noindent{\textbf{#1}}}

\usepackage{listings}
\lstdefinestyle{myStyle}{
      escapeinside={\%*}{*)},
      morekeywords={encoding,
      xs:schema,xs:element,xs:complexType,xs:sequence,xs:attribute}
}

\usepackage{multirow}
\usepackage[table]{xcolor}
\definecolor{altoshade}{gray}{0.92}

\title{ALTO: \underline{A}frican and \underline{L}evantine \underline{T}ech-Facilitated Gender-Based Violence C\underline{o}rpus}

\author{
 \textbf{Tianwei Zhang,\textsuperscript{1}}
 \textbf{Ivy Njoki Nyagwencha,\textsuperscript{1}}
 \textbf{Zahraa Dirani,\textsuperscript{1}}
 \textbf{Shalini Joshi,\textsuperscript{1}}
 \textbf{Scott A. Hale\textsuperscript{1,2}}
\\
 \textsuperscript{1}Meedan,
 \textsuperscript{2}Oxford Internet Institute, University of Oxford,
\\
 \small{
   \textbf{Correspondence:} \href{mailto:terry@meedan.org}{terry@meedan.org}, \href{mailto:scott@meedan.org}{scott@meedan.org}
 }
}

\begin{document}
\maketitle
\begin{abstract}

Tech-facilitated gender based violence (TFGBV) is a pervasive online harm affecting women and gender minority groups through its many direct and covert forms. Current models and annotated datasets for gendered harm detection are dominated by data from, and models trained in, Global North languages. In collaboration with civil society organizations (CSOs) from Jordan, Francophone West Africa, and Kenya, we developed a community-led approach to crowdsource culturally and linguistically relevant instances of TFGBV in Levantine Arabic, African French, and Swahili. We annotated the crowdsourced data using active learning to develop a small but performant dataset ($N$=3022) called ALTO to improve TFGBV detection in the three languages. Through benchmarking and fine-tuning experiments, we show that existing harmful content detection models show poor TFGBV recall within ALTO and improve with fine-tuning. We found that models often miss covert harms across all languages, while error rates are markedly higher in Swahili. We report the experimental results and release the annotation taxonomy, dataset, and trained models.\footnote{Available in \href{https://huggingface.co/collections/meedan/alto}{HuggingFace repository}.}

\end{abstract}

\section{Introduction}
Tech-facilitated gender based violence (TFGBV) is a global problem consistently linked with offline harms, including depression, anxiety, and stress \citep{BenitezHidalgo2026} that affects women and other gender minority populations most acutely \citep{Turner2024}. TFGBV was estimated to impact between 16\% and 58\% of women \citep{hicks_global_2021}, with a disproportionately greater prevalence against women and girls from the Larger World (i.e., Global South) \citep{eiu_measuring_2021}. Online violence also has offline manifestations, with 41\% of women in the public sphere reporting offline violence linked to online encounters in a recent survey \citep{Posetti2025}. %

The use of an automated content moderation system is necessary given the volume of online content \citep{Tonneau2025b}. Two commonly used model architectures for automated TFGBV detection are fine-tuned, encoder-only models such as BERT \citep{Devlin2019} and zero- or few-shot decoder-only models such as OpenAI’s GPT safety fine-tuned models \citep{openai_introducing_2025}. The underlying pre-training and fine-tuning are based on large corpora, often derived from publicly available online text, including social media \citep{luccioni_whats_2021}. Most online content is in English \cite{hale2014global}, and performance degrades for less-represented languages in real-life settings \citep{Tonneau2025a}. Language availability also varies across annotated TFGBV datasets, with English accounting for 41.3\% of available resources, while the next most represented languages, Spanish and Arabic, each account for only 12.7\% \citep[]{abercrombie_resources_2023}. The language disparity in TFGBV detection resources can expose women, girls, and gender minority people in the Larger World to more harm. In this work, we make three major contributions:
\begin{itemize}\vspace{-1ex}
  \setlength{\itemsep}{0pt}
  \setlength{\parskip}{0pt}
  \setlength{\parsep}{0pt}
  \setlength{\topsep}{0pt}
    \item We survey available TFGBV datasets and classifiers for Levantine Arabic, African French, and Swahili, finding that resources in these languages are sparse or form an extremely small part of larger datasets,
    \item We find existing models like Perspective API, GPT-OSS-Safeguard, and Learning from the Worst \citep{Vidgen2021} perform poorly on detecting TFGBV in the three languages with distinct misclassification patterns, and
    \item We develop novel annotated TFGBV datasets using active learning on crowdsourced data and show that they improve model performance. 
\end{itemize}

\section{Related Work}
\smallskip
\para{TFGBV Framework.} This paper operationalizes the \citet{UNFPA2021} TFGBV framework, which defines TFGBV under three aspects: tech facilitation, gender-based, and the presence harm or violence in many forms. Past work has focused narrowly on the conceptualization of misogyny and sexism, whereas TFGBV encompasses various forms of violence perpetrated against individuals on the basis of gender \citep{abercrombie_resources_2023}. \citet{jiang_re-examining_2024} operationalized the TFGBV framework by reannotating past datasets from \citet{kirk_semeval-2023_2023} and \citet{guest_expert_2021}, but their effort was limited by the misogyny and sexism target labels in the underlying datasets. Little academic work has been done since to collect and annotate harmful content using the TFGBV framework. %

\para{Identity Harm Detection.} There is an established line of work on harmful and abusive content detection with labels for target identities \citep[e.g.][]{kwarteng_misogynoir_2022, Tonneau2025a, Ousidhoum2019, yadav_lahm_2023, Muhammad2025}. There is also extensive work on misogynistic and sexist content detection \citep[e.g.][]{mulki_let-mi_2021, guest_expert_2021, kirk_semeval-2023_2023}. Within the past work, we found few resources that adopted TFGBV harm types defined in the latest international guidelines \citep{dunn_technology-facilitated_2020, UNWomen2024}, such as threats and incitement of physical and sexual violence and doxxing \citep{mulki_let-mi_2021, nobata_abusive_2016}. Despite \citeposs{cook_exploring_2009} call to include more harm types, such as physical and sexual violence, when assessing aggression against women, labelling efforts targeting additional harm types remain scarce. We address this gap by developing a fine-grained TFGBV taxonomy that covers a broader range of harm types.

\para{Resources in African and Levantine Languages.} Hate and abusive content detection datasets remain biased toward English and European languages \citep{tonneau-etal-2024-languages}. Arabic is a notable exception with several past datasets developed for harmful content and gendered harm \citep{abercrombie_resources_2023}. However, \citet{haj_ahmed_navigating_2025} documented a dialectal bias toward Modern Standard Arabic, Gulf, and Egyptian varieties, with detection performance degrading on Levantine Arabic. Let-Mi \citep{mulki_let-mi_2021} is the only dataset specifically targeting Levantine Arabic misogyny, while MLMA and ArMIS contain mixed dialects \citep{Ousidhoum2019, almanea_armis_2022}.

French spoken and used online in Africa is strongly influenced by the local languages, forming distinct varieties of French from the European counterpart \citep{boutin_view_2018}. Despite French's status as a high-resource language within NLP research, it remains biased towards Standard French \citep{ahia_low-resource_2021}. Past datasets on French hate speech and gender harm detection \citep[e.g.][]{Ousidhoum2019, chiril2020} also do not represent African varieties, demonstrating a gap in the availability of data for this populous region.

The recent AfriHate corpus \citep{Muhammad2025} contains 15 languages, including Swahili and North African Arabic, and labels instances of hate and targets. Fine-tuned models achieved strong performance in low-resource multilingual hate detection (Swahili F1=0.895). However, at the time of this study, the released AfriHate dataset did not contain the identity target labels for Swahili. Hate speech detection datasets for other African regions have been developed, such as NaijaHate \citep{tonneau_naijahate_2024} and EkoHate \citep{ilevbare_ekohate_2024} for Nigerian languages, \citep{ayele_exploring_2023} and \citep{ibrahim_resources_2024} for Amharic. 

The paucity of gendered harm detection datasets in Levantine Arabic, African French, and Swahili, used by nearly 500 million people,\footnote{Levantine Arabic - 60 million (population of the Levant from Wikipedia), African French - 228 million \citep{oif_langue_2026}, and Swahili 200 million \citep{unesco_world_2021}} forms a central motivation for this work. 

\para{Active Learning (AL).} Despite the promising few-shot performance of LLMs \citep{brown_language_2020}, the social implications of TFGBV detection necessitate that the dataset be annotated through participatory design with annotators knowledgeable in GBV to reflect survivor and feminist perspectives \citep{muller_designing_2021}. By using a model's internal uncertainty to query subsequent annotation candidates, active learning \citep[AL,][]{lewis_sequential_1994} is an established strategy for reducing annotation costs and the potential harm to annotators through fewer annotations. \citet{schroder_revisiting_2022} and \citet{zhang_survey_2022} describe the recent development of uncertainty and diversity sampling techniques in transformer models, which we implement. AL has been applied to text classification in two approaches: through embedding-space diversity \citep{zhang_active_2017} and predictive uncertainty \citep{wang_deep_2024}. We combined both in our implementation (\S\ref{para:active_learning}).

\section{Dataset Creation}
\subsection{Data Collection}
We used an ensemble approach that combined crowdsourcing via community tiplines and social media scraping to establish the data pool. CSO partners operated and promoted the tiplines in their respective regions and languages (Appendix \ref{app:annotation}). The tiplines were operated using Meedan's Check software,\footnote{https://meedan.org/check} and were accessible to community members via WhatsApp, Telegram, and Facebook Messenger. The community tipline enabled individuals to share potential TFGBV content they encountered online. 

While the tipline supported sampling local, community-reported instances, we saw uneven engagement across the regions. Following prior work, we boosted the number of potentially positive instances by using keyword sampling on social media \citep[e.g.][]{Vidgen2021, Kirk2022, Muhammad2025}. The keywords were curated by the CSO partners, based on their past work and existing literature on harmful content in the relevant languages.\footnote{Available in project repository.}

We used the keywords to sample instances from X, where harmful content increased after the ownership change and rebranding to X from Twitter~\citep{Hickey2023, Hickey2025}. Keywords were entered manually into the search bar, and new results were loaded by scrolling while a browser script saved the loaded Tweets to a local file. All identifying user tags and URLs were replaced with a special token to protect user privacy. In total across tiplines and social media, we collected n=8,710 items for Arabic, n=5,628 for African French, and n=4,798 for Swahili, forming the full data pool, where a subset was sampled using AL for targeted annotation.

\begin{table}[t]
\centering
\small
\setlength{\tabcolsep}{3pt}
\begin{tabularx}{\columnwidth}{lrrrl}
\toprule
\textbf{Batch} &
\textbf{ar} &
\textbf{fr} &
\textbf{sw} &
\textbf{Sampling Strategy} \\
\midrule
$\mathcal{B}_1$  & 50  & 100 & 50 & KMeans diversity \\
$\mathcal{B}_2$  & 130 & 100 & 137 & KNN uncertainty and diversity\\
$\mathcal{B}_3$  & 217 & 194 & 113 & PLM entropy+activation: binary \\
$\mathcal{B}_{4+}$  & 557 & 632 & 742 & PLM entropy+activation: multilabel\\
\midrule
Total & 954 & 1026 & 1042 & \\
\bottomrule
\end{tabularx}
\caption{Annotation batch sizes and sampling strategies for Arabic (ar), French (fr), and Swahili (sw). Batches B2--B4+ use a cumulatively growing annotated seed set from previous batches.}
\label{tab:annotation_rounds}
\end{table}                

\subsection{Active Learning}\label{para:active_learning}
We implement AL using a pool-based approach, in which an unlabelled pool $\mathcal{U}$ is queried at each round $t$ for labelling by the annotators, forming the batch $\mathcal{B}_t$. A combination of diversity and uncertainty sampling approaches was used to produce each batch.\footnote{diversity sampling is also called discriminative sampling \citep[e.g.][]{zhang_survey_2022}} Table \ref{tab:annotation_rounds} shows the sampling strategy used for each AL round. Two approaches were combined to maximize uncertainty and diversity as suggested by \citet{monarch_human---loop_2021}: prediction uncertainty was implemented using normalized entropy \citep{shannon1948} and diversity sampling was implemented using model activation uncertainty \citep{zhang_active_2017} (Appendix \ref{app:al_details}). 

\section{Annotation}
\subsection{Recruitment and Training}
\para{Annotators.} To build a dataset for detecting TFGBV, which overwhelmingly targets women and girls \citep{hicks_global_2021}, we recruited annotators who self-identify as women. Annotations from the targeted group often diverge from majority judgements, and standard majority aggregation can obscure their perspectives \citep{fleisig_when_2023}. Prior work has therefore recommended specialized rater pools for content involving identity-based attacks \citep{goyal_is_2022}. Our design follows \citet{kirk_semeval-2023_2023}, who recruited trained annotators who ``all self-identify as women’’ to mitigate labelling bias and include the communities most affected. We acknowledge that a single-gender pool does not capture all perspectives on TFGBV (see Limitations).

Annotators were native speakers from the study regions who were members of or consultants engaged by the CSO partners, consistent with evidence that local expertise is critical for low-resource hate speech annotation \citep{nkemelu_tackling_2023}. Two annotators and one expert annotator were assigned to each language, with the expert responsible for adjudicating disagreements. The expert annotators had previous experience with gendered disinformation research and collaborated closely with the lead researcher during adjudication. Prior to annotation, the lead researcher and expert annotators conducted training sessions for all annotators. We provided annotators with annotation guidelines that included detailed explanations of the annotation process, taxonomy, and worked examples.
\para{Annotator wellbeing.} Exposure to harmful content during annotation can cause vicarious trauma among annotators \citep{dubberley2015making}. Following guidelines from \citet{vidgen_challenges_2019}, we provided annotators with trauma-mitigation training before they began annotating. A content moderation researcher hosted the training sessions and introduced trauma recognition, harm-minimizing approaches to annotating, and techniques for maintaining well-being. Annotation began only after completion of this training. 

\subsection{Labels and Taxonomy}
The annotation labels reflect the \citet{UNFPA2021} TFGBV framework, which we operationalized as three components: tech-facilitation, gendered content, and harm. Given the online nature of the data, tech-facilitation was assumed and not explicitly annotated. Gendered content was defined as a binary label indicating whether the content was directed at or about women, girls, and other vulnerable gender or sexual identities, without consideration of the harm contained. This definition enabled annotating gendered references that are present but harmless, giving the trained models more nuance. Each instance was then labelled for harm and violence types.

We developed a hierarchical taxonomy of harm and violence types based on international civil society and agency frameworks \citep[i.e.][]{dunn_technology-facilitated_2020, UNFPA2021, UNWomen2024} and prior work on hate and harmful content detection \citep{Bianchi2022,cook_exploring_2009, kirk_semeval-2023_2023, MulkiGhanem2021}. As international guidelines aim to capture the broad range of violence encompassed by TFGBV, they often lack clear, fine-grained class definitions that have been shown necessary for hate speech annotation in machine learning applications \citep{Banko2020, VidgenDerczynski2020}. The resulting taxonomy refines high-level TFGBV types (Tier 1) into fine-grained subtypes (Tier 2), with clear annotation instructions and examples for each subtype (Appendix \ref{app:annotation}).

\subsection{Annotation Procedure} 
The annotation was completed using the HumanSignal platform.\footnote {https://www.humansignal.com} We followed the prescriptive paradigm of annotation, in which annotators applied the labels and taxonomy based on researcher-defined labels \citep{rottger_two_2022, cabitza_toward_2023}. While the prescriptive paradigm discourages subjective label interpretations, we adopted a participatory design approach that incorporated annotator feedback.

After each of the two initial batches of annotation ($\mathcal{B}_1$ \& $\mathcal{B}_2$), we met with the annotators to discuss disagreements. This process revealed limitations in the annotation guidelines and differing interpretations, leading to refinements for subsequent batches. For example, we found that the initial definition of gendered content was not adequately defined for instances where men were targeted using female gendered slurs. The discussion helped clarify that such content is gendered but not TFGBV in most cases, as such content does not directly target a minoritised gender. An exception was agreed for egregious use of gendered slurs where such use is likely distressful to readers in the target group. After the annotation guideline change, previous instances were re-annotated under the new guidelines in subsequent rounds and the latest annotation was kept to ensure coherency. 

\subsection{Annotation Results}
\begin{table}
\small
\centering
\begin{tabularx}{\columnwidth}{Xlrrr}
\toprule
\textbf{Variable} & \textbf{Label} & \textbf{ar} & \textbf{fr} & \textbf{sw} \\
\midrule
\multirow{2}{*}{Gender} & Non-Gendered & 222 & 236 & 395 \\
 & Gendered & 486 & 746 & 439 \\
\midrule
\multirow{6}{*}{Tier 1} & Non-TFGBV & 446 & 525 & 385 \\
 & Doxxing & 0 & 0 & 87 \\
 & Harass. \& Hate Speech & 248 & 382 & 268 \\
 & Threats \& Incite. & 32 & 24 & 12 \\
 & Unsure & 6 & 1 & 1 \\
\midrule
\multirow{10}{*}{Tier 2} & Animosity & 32 & 94 & 27 \\
 & Damning & 20 & 1 & 0 \\
 & Derogation & 107 & 38 & 133 \\
 & Doxx. Threats & 0 & 0 & 64 \\
 & Doxx. Disclosure & 0 & 0 & 21 \\
 & Obj. and Dehuman. & 28 & 4 & 20 \\
 & Sexual Harass. & 11 & 0 & 6 \\
 & Threats \& Incite. Phys. & 8 & 19 & 3 \\
 & Threats \& Incite. Sex & 3 & 4 & 4 \\
\midrule
Total & & 954 & 1026 & 1042 \\
\bottomrule
\end{tabularx}
\caption{Counts of adjudicated or fully agreed labels for Arabic (ar), French (fr), and Swahili (sw) per variable. Total represents the number of annotated instances, differing from the variable-level sums.}
\label{tab:annotation_result}
\end{table}

The distribution of labels is shown in  Table \ref{tab:annotation_result}. Gendered content was more frequent than non-gendered content in all languages, indicating a substantial presence of gendered expression in our samples. In Tier 1 of our taxonomy, \textit{Harassment and Hate Speech} was the most prevalent TFGBV category, with 248 instances in Arabic, 382 in French, and 268 in Swahili. There were the most instances of \textit{Threats and Incitements of Violence} in Arabic than the other languages. By contrast, Swahili had significantly more instances of \textit{Doxxing}, with none in our French or Arabic data. We note, however, that our datasets are not designed to be representative of content online.

For the finer-grained Tier 2 of our taxonomy, we noted an imbalance across languages. \textit{Derogation} was the most common label within \textit{Harassment and Hate Speech} for Arabic and Swahili, with 107 and 133 instances, respectively, while animosity was the most common in French with 94. The Arabic dataset contained 22 instances of Damning, a unique Arabic label introduced by \citet{mulki_let-mi_2021}, one in French and none in Swahili.  Overall, the annotation results indicate differences in how TFGBV manifests across languages and the ability of our sampling approach to capture positive TFGBV instances.

\begin{table}[h]
\centering
\small
\begin{tabularx}{\columnwidth}{l *{3}{>{\centering\arraybackslash}X}}
\toprule

\textbf{Lang.} & \textbf{Gender $\alpha$} & \textbf{Tier-1 $\alpha$} & \textbf{Tier-2 $\alpha$} \\
\midrule
\textbf{ar} & 0.399 & 0.538 & 0.627 \\
\textbf{fr} & 0.661 & 0.534 & 0.451 \\
\textbf{sw} & 0.553 & 0.474 & 0.624 \\
\bottomrule
\end{tabularx}
\caption{Inter-annotator agreement (Krippendorff's $\alpha$) for Arabic (ar), French (fr) and Swahili (sw) across annotation variables.}
\label{tab:iaa_pooled}
\end{table}

The inter-annotator agreement was measured for each annotated variable (Table \ref{tab:iaa_pooled}) using Krippendorff's alpha ($\alpha$). 
For whether content was gendered, we had the lowest agreement for Arabic ($\alpha=0.399$) while French ($\alpha=0.661$) and Swahili ($\alpha=0.553$) were higher. We had moderate agreement for Tier-1 ($\alpha=0.538$) and Tier-2 ($\alpha=0.627$) labels in Arabic

A similar pattern of higher alpha for Tier 2 than Tier 1 was also observed in Swahili, indicating that Tier 2 labelling tends to be more consistent once a Tier 1 label was agreed upon. This was, however, not seen for French, as alpha was lower for Tier 2 than for Tier 1. The moderate agreement in our dataset is broadly consistent with those reported in comparable harmful content annotation tasks. For example, \citet{lee_exploring_2024} reported $\alpha$ between 0.34 and 0.46 for annotating English hate speech, and \citet{Muhammad2025} reported free marginal Randolph's $\kappa$ between 0.46 and 0.81 for hate speech in 15 African languages. The moderate IAAs across the three variables reflect the subjective and difficult nature of TFGBV annotation.

\section{Experiments}

\begin{table*}[t]
\centering
\small
\setlength{\tabcolsep}{6pt}
\begin{tabular}{l l ccc l c}
\toprule
\textbf{Dataset} & \textbf{Lang.} & \textbf{Gender} & \textbf{Harm} & \textbf{Public} & \textbf{Labels} & \textbf{Incl.} \\
\midrule
AfriHate~\citep{Muhammad2025}          & sw         & \cmark & \cmark & \tmark & Hate with target                 & \xmark$^{a}$ \\
HateDay~\citep{Tonneau2025a}           & ar, sw, fr & \cmark & \cmark & \cmark & Hate with target                 & \cmark \\
LAHM~\citep{yadav_lahm_2023}           & ar, fr     & \xmark & \cmark & \xmark & Fine-grained hate incl.\ sexism  & \xmark \\
ArMIS~\citep{almanea_armis_2022}       & ar         & \tmark & \cmark & \cmark & Misogyny and sexism              & \cmark \\
French Sexism~\citep{chiril2020}       & fr         & \cmark & \tmark & \tmark & Sexism                           & \xmark$^{b}$ \\
ArMi~\citep{mulki_working_2022}        & ar         & \tmark & \cmark & \cmark & Misogyny                         & \cmark \\
MLMA~\citep{Ousidhoum2019}             & ar, fr     & \cmark & \cmark & \cmark & Hate with target                 & \cmark \\
\bottomrule
\end{tabular}
\caption{Candidate TFGBV datasets that cover gender, targets and harm, and whether they are included (Incl.) in the \texttt{Bench} dataset (\cmark~= yes, \xmark~= no, \tmark~= partial). $^{a}$Gender labels not public. $^{b}$Counter-speech annotated as positives}
\label{tab:datasets}
\end{table*}

\subsection{Experimental Setup}
We designed two related experiments to test our dataset's ability to improve TFGBV detection. The first benchmark experiment examines the performance of existing off-the-shelf models on \texttt{ALTO} and existing datasets. In the second, we fine-tune on a portion of the benchmarking data (Bench) and our annotated data (ALTO) and conduct error analysis to understand misclassification patterns. 

\para{Existing Data.} Our survey of datasets identified a number of candidate models, and a set of criteria was applied for inclusion. Our benchmarking data includes all datasets that: 1) are labelled for gender, 2) fit the definition of harm under our TFGBV framework, and 3) are publicly accessible. 

We reviewed 140 datasets from \citet{abercrombie_resources_2023}, \citet{vidgen_directions_2020} and a targeted search on HuggingFace and identified four mono- and multilingual datasets for inclusion. All four support Arabic of mixed dialects, three support French with no specific varieties, and one supports Swahili as a subset of a larger Kenyan dataset, highlighting a gap in TFGBV detection resources for these languages. Furthermore, in large datasets, such as the representative sample from HateDay \citep{Tonneau2025a} with 45,000 items from Kenya, only 17 TFGBV instances were captured, demonstrating the need for targeted sampling. 

\para{Existing models.} We report the performance of existing harm detection models to establish baseline TFGBV detection performance. We identified candidate models from the hate and harm detection literature, commercial solutions, and the open-source model repository HuggingFace. Candidates were included based on criteria that replicate models readily implementable for content moderation by small- to medium-sized platforms. Specifically, we included models based on three criteria: 1) language support, 2) public availability in a pre-trained state without requiring further fine-tuning with a transparent model card or methodology, and 3) architectural diversity, covering both decoder-only generative models and encoder-based classifiers. The benchmarked models included models from OpenAI \citep{openai_introducing_2025}, \citet{Vidgen2021}, and Perspective API \citep{Google2026} (Appendix \ref{app:benchmark_models}). The fine-tuned models are encoder-only models most commonly used in NLP research for their respective languages (Appendix \ref{app:finetune_models}).

\begin{table}
\small
\centering
\begin{tabularx}{\columnwidth}{l *{4}{>{\raggedleft\arraybackslash}X}}
\toprule
\textbf{Lang} & \textbf{ALTO-train} & \textbf{ALTO-test} & \textbf{Bench-train} & \textbf{Bench-test} \\
\midrule
\textbf{ar} & 467 (44.1\%) & 117 (44.4\%) & 3316 (13.3\%) & 829 (13.3\%) \\
\textbf{fr} & 718 (44.4\%) & 180 (45.0\%) & 3632 (2.9\%) & 907 (2.9\%) \\
\textbf{sw} & 508 (38.0\%) & 127 (37.8\%) & 564 (2.5\%)$^{\dagger}$ & 140 (2.1\%)$^{\dagger}$ \\
\midrule
\textbf{Total} & \textbf{1693 (42.4\%)} & \textbf{424 (42.7\%)} & \textbf{7512 (7.4\%)} & \textbf{1876 (7.4\%)} \\
\bottomrule
\end{tabularx}
\caption{Dataset size by language and split, with the positive TFGBV class as a percentage of each split's total (in brackets). $^{\dagger}$Swahili positive-class support in \texttt{Bench }is very low ($n=14$ train, $n=3$ test); metrics on this subset has limited statistical power.}
\label{tab:label-distribution}
\end{table}

\para{Data setup.} \label{para:data_setup}The surveyed datasets were combined into language-specific benchmark datasets (Table \ref{tab:label-distribution}). The labels in the original datasets were standardized to produce binary gendered content and harm labels if they target women and girls. Similarly, we standardized the ALTO labels to binary gendered and harm labels for comparability. For the ALTO dataset, we retained only annotations that achieved full agreement, were adjudicated, or had minor disagreements (i.e., Tier 1 agreed) resulting in $N=2117$ for the experiments but we release the full dataset with all annotations. The benchmark datasets (Bench) and ALTO were split into 80\% for training and 20\% for testing. The ALTO dataset split was stratified on binarized Tier 1, gendered content, and batches. Bench was stratified from each dataset on binarized labels of gendered content and harm label.

\para{Benchmarking experiment.} We evaluated the performance of existing harmful content detection models on the surveyed \texttt{Bench-test} and \texttt{ALTO-test}. All benchmarking experiments were completed using the binarized TFGBV labels. The ALTO model results were reported at 5-fold cross-validation (CV) using out-of-fold performance. The LFTW model was trained only on English data, which does not meet our criteria. However, it is a widely used pre-trained hate speech detection tool, is frequently benchmarked in the harmful content literature, and has demonstrated strong performance \citep[e.g.][]{Bianchi2022, Tonneau2025a}. We addressed LFTW's linguistic limitations by translating the collected data into English using the Google Translate API.\footnote{https://cloud.google.com/translate} Translating harmful content from low-resource languages allowed us to assess English-only models' ability to capture culturally and linguistically specific expressions of TFGBV.

\paragraph{Fine-tuning experiment.} We fine-tuned three monolingual and one multilingual pretrained language models (PLMs) for the three languages. Each PLM was initialized from their released checkpoint with a three-head classification layer fitted for the binarized annotation targets of \textit{gendered content}, \textit{presence of harmful content}, and \textit{TFGBV}. The ALTO fine-tuning was evaluated using 5-fold CV, while  \texttt{Bench-train} was kept constant within the training data. All training hyperparameters are reported in Appendix \ref{app:exp_setup}.

\newcolumntype{Y}{>{\centering\arraybackslash}X} 

\section{Experimental Results}
\begin{table*}[h]
\centering
\small
\setlength{\tabcolsep}{3pt}
\begin{tabularx}{\textwidth}{l *{18}{Y}}
\toprule
 & \multicolumn{6}{c}{\textbf{Arabic (ar)}} & \multicolumn{6}{c}{\textbf{French (fr)}} & \multicolumn{6}{c}{\textbf{Swahili (sw)}} \\
\cmidrule(lr){2-7} \cmidrule(lr){8-13} \cmidrule(lr){14-19}
 & \multicolumn{3}{c}{\texttt{ALTO}} & \multicolumn{3}{c}{\texttt{Bench}} & \multicolumn{3}{c}{\texttt{ALTO}} & \multicolumn{3}{c}{\texttt{Bench}} & \multicolumn{3}{c}{\texttt{ALTO}} & \multicolumn{3}{c}{\texttt{Bench}} \\
\cmidrule(lr){2-4} \cmidrule(lr){5-7} \cmidrule(lr){8-10} \cmidrule(lr){11-13} \cmidrule(lr){14-16} \cmidrule(lr){17-19}
\textbf{Model} & P & R & F1 & P & R & F1 & P & R & F1 & P & R & F1 & P & R & F1 & P & R & F1 \\
\midrule
OSS-SG-120B & \textbf{.890} & .624 & .733 & .371 & .418 & .393 & .633 & .682 & .657 & .200 & .577 & .297 & \textbf{.758} & .390 & .515 & .000 & .000 & .000 \\
OSS-SG-20B  & .825 & .640 & .721 & .362 & .536 & .432 & .626 & .733 & .675 & .165 & .654 & .264 & .595 & .427 & .498 & .000 & .000 & .000 \\
LFTW        & .705 & .667 & .685 & .288 & .427 & .344 & .603 & .635 & .619 & .058 & .654 & .107 & .546 & .469 & .504 & .000 & .000 & .000 \\
P-API       & .747 & .686 & .715 & .127 & .391 & .192 & .564 & .748 & .643 & .041 & \textbf{.769} & .078 & -- & -- & -- & -- & -- & -- \\
\midrule
\rowcolor{altoshade}
ALTO (mono)  & .844 & .795 & \textbf{.818} & \textbf{.561} & .627 & .592 & \textbf{.697} & .787 & \textbf{.739} & \textbf{.359} & .538 & \textbf{.431} & .601 & .780 & \textbf{.679} & \textbf{.118} & \textbf{.667} & \textbf{.200} \\
\rowcolor{altoshade}
ALTO (multi) & .686 & \textbf{.837} & .754 & .479 & \textbf{.936} & \textbf{.634} & .582 & \textbf{.875} & .699 & .189 & .385 & .253 & .573 & \textbf{.817} & .674 & \textbf{.118} & \textbf{.667} & \textbf{.200} \\
\bottomrule
\end{tabularx}
\caption{TFGBV-class precision (P), recall (R), and F1 by language on the full \texttt{ALTO} dataset ($n{=}2{,}117$) and \texttt{Bench-test}. \colorbox{altoshade}{Shaded} rows are encoders fine-tuned on ALTO, and their ALTO values are 5-fold CV predictions (inter-fold $\sigma\le 0.08$). \textit{ALTO (mono)}: monolingual models \textit{ALTO (multi)}: a single multilingual model trained on all three languages. P-API was not evaluated on sw (--). \textbf{Bold} = best per metric.}
\label{tab:main_results}
\end{table*}
\subsection{Benchmarking Results}
Table \ref{tab:main_results} shows the benchmarking results on the positive TFGBV class. We find that evaluating on ALTO, existing models have language-dependent magnitudes of F1 ranges. Within the benchmarked models, Arabic has the highest F1 between 0.685 and 0.733, African French is lower at 0.619--0.675, and Swahili is the lowest at 0.498--0.504. 

We observed a drop in performance across both existing and fine-tuned models when changing the evaluation set from \texttt{ALTO} to \texttt{Bench}. The F1 scores of benchmarked models dropped from a simple average of 0.675 on \texttt{ALTO} to an average of 0.202 on \texttt{Bench}. Recall between evaluation sets regressed less from an average recall of 0.682 for \texttt{ALTO} to 0.485 for \texttt{Bench} (Appendix \ref{app:benchmark_results}). The extreme positive label bias in \texttt{Bench} is especially evident for Swahili where none of the 3 positive samples were detected the off-the-shelf models.  

The best and worst models for each language differed. Across all languages on \texttt{ALTO}, OSS-SG-20B and 120B were the best off-the-shelf models. Interestingly, the 20B model often achieved better recall of TFGBV than the larger 120B counterpart while the latter achieved better precision. The worst-performing model for TFGBV detection on ALTO was the LFTW model which relied on translated texts with the exception of Swahili where OSS-SG-20B performed slightly worst.

\subsection{Fine-tuning Results}
Fine-tuning on \texttt{ALTO} and \texttt{Bench} significantly improved TFGBV detection performance beyond existing models, primarily by improving recall (Appendix \ref{app:significance_test}). Arabic and French improved from the best recall of 0.686 and 0.748 from Perspective API to 0.837 and 0.875 by the multilingual ALTO model after fine-tuning. Swahili recall improved from 0.469 by LFTW to 0.817 by the multilingual model. Precision improved modestly on \texttt{ALTO} evaluation, while a larger improvement was observed on \texttt{Bench} evaluation. Monolingual fine-tuning achieved higher F1 for all languages across \texttt{ALTO} and \texttt{Bench} except for Arabic \texttt{Bench} where the multilingual model excelled. In general, multilingual fine-tuning achieved higher recall but worse precision than the monolingual models.

\subsection{Error Analysis}
\begin{table*}[h]
\centering
\small
\setlength{\tabcolsep}{2.5pt}
\renewcommand{\arraystretch}{0.9}
\begin{tabular}{c l r cccccc}
\toprule
 & TFGBV Tier-2 Category & $n$ & OSS-SG-120B & OSS-SG-20B & LFTW & P-API & ALTO (mono) & ALTO (multi) \\
\midrule
\multirow{7}{*}{\rotatebox[origin=c]{90}{\textbf{Arabic}}}
 & Derogation & 103 & \cellcolor{red!31}.35 & \cellcolor{red!33}.38 & \cellcolor{red!26}.30 & \cellcolor{red!23}.26 & \cellcolor{red!12}.14 & \cellcolor{red!9}.11 \\
 & Animosity & 32 & \cellcolor{red!44}\textcolor{white}{.50} & \cellcolor{red!44}\textcolor{white}{.50} & \cellcolor{red!52}\textcolor{white}{.59} & \cellcolor{red!57}\textcolor{white}{.66} & \cellcolor{red!41}\textcolor{white}{.47} & \cellcolor{red!19}.22 \\
 & Objectification & 26 & \cellcolor{red!20}.23 & \cellcolor{red!20}.23 & \cellcolor{red!24}.27 & \cellcolor{red!13}.15 & \cellcolor{red!13}.15 & \cellcolor{red!7}.08 \\
 & Damning & 16 & \cellcolor{red!44}\textcolor{white}{.50} & \cellcolor{red!38}\textcolor{white}{.44} & \cellcolor{red!22}.25 & \cellcolor{red!27}.31 & \cellcolor{red!27}.31 & \cellcolor{red!33}{.38} \\
 & Sexual harassment & 10 & \cellcolor{red!0}.00 & \cellcolor{red!9}.10 & \cellcolor{red!0}.00 & \cellcolor{red!9}.10 & \cellcolor{red!0}.00 & \cellcolor{red!0}.00 \\
 & Non-TFGBV, gendered & 136 & \cellcolor{blue!10}.12 & \cellcolor{blue!16}.18 & \cellcolor{blue!18}.21 & \cellcolor{blue!15}.17 & \cellcolor{blue!20}.23 & \cellcolor{blue!51}\textcolor{white}{.59} \\
 & Non-TFGBV, not gendered & 190 & \cellcolor{blue!2}.02 & \cellcolor{blue!5}.05 & \cellcolor{blue!20}.23 & \cellcolor{blue!17}.19 & \cellcolor{blue!3}.04 & \cellcolor{blue!9}.10 \\
\midrule
\multirow{5}{*}{\rotatebox[origin=c]{90}{\textbf{French}}}
 & Animosity & 91 & \cellcolor{red!40}\textcolor{white}{{.46}} & \cellcolor{red!35}{.40} & \cellcolor{red!42}\textcolor{white}{.48} & \cellcolor{red!41}\textcolor{white}{{.47}} & \cellcolor{red!26}.30 & \cellcolor{red!15}.18 \\
 & Derogation & 38 & \cellcolor{red!39}\textcolor{white}{.45} & \cellcolor{red!32}.37 & \cellcolor{red!46}\textcolor{white}{{.53}} & \cellcolor{red!18}.21 & \cellcolor{red!28}.32 & \cellcolor{red!14}.16 \\
 & Threats of violence & 17 & \cellcolor{red!15}.18 & \cellcolor{red!10}.12 & \cellcolor{red!21}.24 & \cellcolor{red!10}.12 & \cellcolor{red!31}{.35} & \cellcolor{red!15}{.18} \\
 & Non-TFGBV, gendered & 278 & \cellcolor{blue!48}\textcolor{white}{.54} & \cellcolor{blue!51}\textcolor{white}{.59} & \cellcolor{blue!41}\textcolor{white}{.46} & \cellcolor{blue!53}\textcolor{white}{.60} & \cellcolor{blue!39}\textcolor{white}{.45} & \cellcolor{blue!69}\textcolor{white}{.79} \\
 & Non-TFGBV, not gendered & 220 & \cellcolor{blue!3}.03 & \cellcolor{blue!5}.05 & \cellcolor{blue!15}.17 & \cellcolor{blue!25}.29 & \cellcolor{blue!5}.06 & \cellcolor{blue!12}.14 \\
\midrule
\multirow{6}{*}{\rotatebox[origin=c]{90}{\textbf{Swahili}}}
 & Derogation & 120 & \cellcolor{red!49}\textcolor{white}{.56} & \cellcolor{red!48}\textcolor{white}{.55} & \cellcolor{red!39}\textcolor{white}{.44} & -- & \cellcolor{red!18}.20 & \cellcolor{red!13}.15 \\
 & Animosity & 21 & \cellcolor{red!54}\textcolor{white}{.62} & \cellcolor{red!46}\textcolor{white}{.52} & \cellcolor{red!29}.33 & -- & \cellcolor{red!12}.14 & \cellcolor{red!0}.00 \\
 & Objectification & 20 & \cellcolor{red!57}\textcolor{white}{{.65}} & \cellcolor{red!57}\textcolor{white}{.65} & \cellcolor{red!70}\textcolor{white}{.80} & -- & \cellcolor{red!39}\textcolor{white}{.45} & \cellcolor{red!35}{.40} \\
 & Doxxing & 13 & \cellcolor{red!54}\textcolor{white}{.62} & \cellcolor{red!67}\textcolor{white}{{.77}} & \cellcolor{red!67}\textcolor{white}{.77} & -- & \cellcolor{red!40}\textcolor{white}{{.46}} & \cellcolor{red!27}.31 \\
 & Non-TFGBV, gendered & 89 & \cellcolor{blue!17}.19 & \cellcolor{blue!35}\textcolor{white}{.40} & \cellcolor{blue!29}.33 & -- & \cellcolor{blue!58}\textcolor{white}{.66} & \cellcolor{blue!66}\textcolor{white}{.75} \\
 & Non-TFGBV, not gendered & 305 & \cellcolor{blue!4}.04 & \cellcolor{blue!10}.11 & \cellcolor{blue!19}.21 & -- & \cellcolor{blue!19}.22 & \cellcolor{blue!23}.26 \\
\bottomrule
\end{tabular}
\caption{Error rate per tier-2 category where available and model on \texttt{ALTO-full} (5-fold out-of-fold prediction). \colorbox{red!35}{Red} rows: FNR of TFGBV positives. \colorbox{blue!35}{Blue} rows: FPR on the non-TFGBV items split by presence of gendered content. Darker shading means more errors and lower error rates are better for all rows.}
\label{tab:fn_category}
\end{table*}

Table \ref{tab:fn_category} shows the errors by TFGBV subcategory measured by false-negative rates (FNR) and false-positive rates (FPR). \textit{Animosity}, often a more covert form of harm, is the category most misclassified in nearly all languages. For this category, models showed an FNR between 0.33 and 0.66 across all languages. Fine-tuning on ALTO decreased FNR, and multilingual fine-tuning improved it substantially (0.00--0.22) over monolingual fine-tuning (0.14--0.47). The more direct insult type, \textit{derogation}, showed a lower FNR range across languages: 0.26--0.38 for Arabic, 0.21--0.53 for French, and 0.33--0.62 for Swahili. 

Between languages, Swahili showed a higher FNR range for all labels (0.33--0.80) than Arabic (0.00--0.66) and French (0.12--0.47). In Swahili, existing models struggled with objectification and doxxing, with OSS-SG between 0.62--0.77 and LFTW highest at 0.77--0.80. The high error rate for LFTW shows that translated text does not meaningfully convey harmful signals in Swahili. Fine-tuning on ALTO, even with only tens of labelled instances, improves doxxing FNR by -0.31 and objectification FNR by -0.25 from OSS-SG-120B to multilingual. 

\section{Discussion}
\paragraph{Regression between datasets.} The reduction of F1 for all models between \texttt{ALTO} and \texttt{Bench} reported in Table \ref{tab:main_results} were mostly driven by lower precision. In the \texttt{Bench} dataset, the proportion of positive TFGBV samples is markedly smaller (7\%) than in the \texttt{ALTO} test set (42.7\%, Table \ref{tab:label-distribution}). The reduction in model performance when evaluating on a real-world dataset was also observed by previous work \citep{Tonneau2025a}, where model precision averaged 0.094 on realistic datasets against 0.40 on academic datasets. Including \texttt{Bench} thus shows realistic performance, which ALTO fine-tuning improves.

\paragraph{Language resource performance.} Within NLP research, Arabic and French are considered high-resource languages despite our studied dialects being lower-resourced \citep{ahia_low-resource_2021}. Our fine-tuning experiment showed that these two languages adapted well to new dialectal data and achieved strong final performance, given their larger pre-training resources. By contrast, Swahili saw the largest absolute gains (F1 + 0.17-0.18, Appendix \ref{app:significance_test}) but only modest final performance for comparable training data size, suggesting that improving model pre-training resources is necessary for domain-specific adaptations such as TFGBV detection to produce larger gains \citep{wu_are_2020}.

\paragraph{Fine-tuning tradeoffs.} While multilingual ALTO-fine-tuned models achieved the lowest FNR across nearly all categories, except Arabic \textit{Damning} and French \textit{Threats}, we observed a trade-off: significantly higher FPR for negative, gendered items, including items that discuss gendered content or use a gendered word. For example, the fine-tuned models predicted the phrase \textbf{fr:}``\textit{Il m’as }[sic]\textit{ clairement pris pour une go facile. Mais bon il s’est tiré une balle dans le pied là}" (\textbf{en:} he clearly took me as an easy girl but he shot himself in the foot.) was flagged because ``go facile" is an African French slang for a promiscuous girl that is used in a reclaimed manner and is a observed failure mode \citep{rottger_hatecheck_2021}. The higher FPR in multilingual models than in monolingual models for gendered negative samples could reflect a loss of language-specific nuance from multilingual pre-training, as suggested by \citet{chang_when_2024}.

\section{Conclusion}
TFGBV is a serious and prevalent harm facing women and gender minorities online that disproportionately affects those in regions with fewer detection resources. To address this gap, we introduced ALTO, a novel annotated dataset for TFGBV detection in Levantine Arabic, African French, and Swahili. We used active learning to sample annotations from both community-reported and scraped sources. We find that ALTO fine-tuning improves TFGBV detection, whereas off-the-shelf harmful content detection models do not generalize well to specific TFGBV subtypes in these languages. Our work showed that targeted, community-led data collection and active learning can improve model performance with modest data sizes for low-resource settings.

\section*{Limitations}
There are several limitations of this work that may be opportunities for future works. The first is the use of binarized labels mapped from the original multi-label annotation in the experiments. The binary labels for TFGBV compress the nuance in Tier 1 and Tier 2 TFGBV, improving prediction performance but reducing interpretability. The decision was motivated by the small number of agreed instances for the Tier 1 and 2 labels, a similar challenge and approach that were reported by \citet{Bianchi2022}. Nonetheless, we release the dataset with all agreed or adjudicated labels (i.e., gendered content, Tier 1 and 2 TFGBV) for other researchers to use.

The data collected from social media in this study were limited to one platform, X. This aligns with many previous studies but limits the potential manifestations of TFGBV to those unmoderated on X. The tipline submissions included various sources, but we excluded those from image- or video-based platforms (e.g., Instagram, TikTok), as this paper focused on text-based TFGBV. Future work can sample potential TFGBV from more platforms and media types. We also hope to pursue future work using the diverse forms of media omitted in this study. 

There is also a linguistic limitation to the data collection approach. While we used keywords from the Levantine and African French contexts, the social media sampling approach could not guarantee that the returned samples were exclusively Levantine Arabic and African French. However, because the annotators are from the respective regions, the annotation approach should capture Levantine and Francophone African perspectives on TFGBV. For Swahili, the common usage of Sheng, a mix of Swahili and English, also means that our samples are not purely Swahili. We believe this linguistically nuanced sample more accurately reflects platform users in these regions. 

We also recognize a limitation in the annotator demographics of this study. While TFGBV targets women, it is also prevalent against other minority sexual and gender identities \citep{UNFPA2021}. All of the annotators in this study were women, which limits the perspectives of other minority identities, such as LGBTQ+ individuals, who can also be survivors of TFGBV \citep{cabitza_toward_2023}. A considerable body of work has shown that annotator identities and beliefs can strongly influence labelling and classifier behaviours \citep[e.g.][]{fleisig_when_2023, jiang_re-examining_2024, waseem_are_2016}, we encourage future work to include annotators from diverse identities without diminishing the perspective of the target identity annotators.

We cannot fully rule out \textit{data contamination} in our experiments \citep{sainz_nlp_2023}, since the benchmark datasets and models also contained Twitter data. While we limited our Twitter/X sampling to content from the last two years (i.e. after most benchmarked resources' time periods), this does not guarantee a mutually exclusive sample. Despite that, the cross-validation on held-out data in \texttt{ALTO} and \texttt{Bench}, with varying and moderately strong performance, suggests that data contamination in ALTO may be limited. 

Lastly, we did not evaluate the English translation used to evaluate the LFTW model at scale because the translation classification experiment was designed to demonstrate TFGBV detection performance when using a publicly available translation service. Therefore, translation-quality effects on prediction are part of the experimental design rather than a confounding factor. We do note that in the N=20 translations from each language that annotators reviewed, recent and covert TFGBV nuances are often missed by machine translations. 

\section*{Ethical Considerations}
\para{Risks.} The dataset contains harmful content that may be distressing to users. We include a content warning before the harmful sections of the paper and note the same for the research repository. There is also the risk of the data being used for malicious purposes, such as training harmful generative models, we prohibit this under the Intended Use.

\para{Privacy.} To protect the identity of individuals whose data we have included, all user tags and URLs were replaced with the respective representative token <USER> and <URL>. Additionally, content labelled as \textit{Doxxing} was excluded from the dataset to protect the survivors. 

\para{Intended Use.} Our dataset is released under a CC BY license and requires prospective users to agree to use the data consistent with its intended purpose before they are able to download it.\footnote{\url{https://creativecommons.org/licenses/by/4.0//}}

We further \textbf{strictly prohibit} the use of ALTO to train, fine-tune, or develop harmful generative models. Use of the dataset in developing content moderation systems should be in conjunction with a human moderation process.

\para{Past Datasets and Models.} All past datasets and models were released under CC BY 4.0 and Apache 2.0 licenses, which our use for research is permitted. Perspective API's use for research falls within Google API Terms of Service\footnote{\url{https://developers.google.com/terms/}}. 

\section*{Acknowledgments}

This work would not have been possible without the continuous contributions of the annotators from the Kenyan ICT Action Network (KICTANet), the Jordan Open Source Association (JOSA), and the Brain Builders Youth Development Initiative. This work was supported by the Swedish International Development Cooperation Agency (Sida), the United Nations Population Fund (UNFPA), and the Christchurch Call Foundation as part of the Project Catalyst consortium, in which peer inputs from the consortium supported conceptualization and methodology development. Hale also acknowledges the support of the ESRC Digital Good Network (ES/X502352/1). We would further like to thank UNFPA for their contributions to survivor data protection and handling.
The views expressed are those of the authors and do not necessarily reflect the views of any of the supporting organizations or collaborators.

\bibliography{custom}

\clearpage

\newpage
\appendix
\onecolumn
\section{Additional Results}
\subsection{Classification Performance}
\label{app:roc_curve}
\begin{figure*}[h!]
  \includegraphics[width=\linewidth]{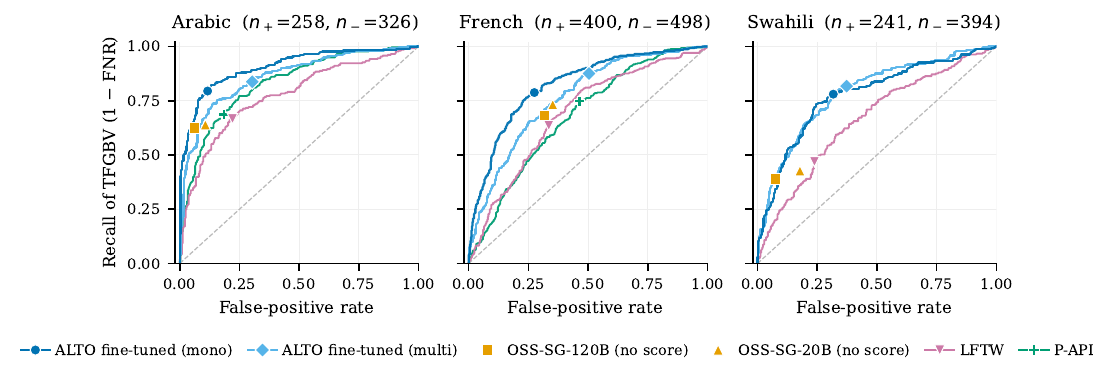}
  \caption{ROC for full ALTO over 5-fold CV and benchmarked models.}
  \label{fig:roc_curve}
\end{figure*}

\subsection{Full Benchmarking Results} \label{app:benchmark_results}
\begin{table*}[h!]
\centering
\small
\setlength{\tabcolsep}{4pt}
\resizebox{\textwidth}{!}{%
\begin{tabular}{l l ccc ccc @{\hskip 10pt} ccc ccc}
\toprule
 & & \multicolumn{6}{c}{\texttt{\textbf{ALTO-full}}} & \multicolumn{6}{c}{\texttt{\textbf{Bench-test}}} \\
\cmidrule(lr){3-8} \cmidrule(lr){9-14}
 & & \multicolumn{3}{c}{\textbf{Non-TFGBV}} & \multicolumn{3}{c}{\textbf{TFGBV}} & \multicolumn{3}{c}{\textbf{Non-TFGBV}} & \multicolumn{3}{c}{\textbf{TFGBV}} \\
\cmidrule(lr){3-5} \cmidrule(lr){6-8} \cmidrule(lr){9-11} \cmidrule(lr){12-14}
\textbf{Lang} & \textbf{Model} & \textbf{P} & \textbf{R} & \textbf{F1} & \textbf{P} & \textbf{R} & \textbf{F1} & \textbf{P} & \textbf{R} & \textbf{F1} & \textbf{P} & \textbf{R} & \textbf{F1} \\
\midrule
\multirow{6}{*}{ar}
 & OSS-SG-120B & 0.759 & \textbf{0.939} & 0.840 & \textbf{0.890} & 0.624 & 0.733 & 0.909 & 0.892 & 0.900 & 0.371 & 0.418 & 0.393 \\
 & OSS-SG-20B & 0.758 & 0.893 & 0.820 & 0.825 & 0.640 & 0.721 & 0.923 & 0.855 & 0.888 & 0.362 & 0.536 & 0.432 \\
 & LFTW & 0.747 & 0.779 & 0.763 & 0.705 & 0.667 & 0.685 & 0.905 & 0.839 & 0.871 & 0.288 & 0.427 & 0.344 \\
 & P-API & 0.767 & 0.816 & 0.790 & 0.747 & 0.686 & 0.715 & 0.863 & 0.588 & 0.700 & 0.127 & 0.391 & 0.192 \\
\rowcolor{altoshade}
 \cellcolor{white} & MARBERTv2 & \textbf{0.845} & 0.883 & \textbf{0.864} & 0.844 & 0.795 & \textbf{0.818} & 0.942 & \textbf{0.925} & \textbf{0.933} & \textbf{0.561} & 0.627 & 0.592 \\
\rowcolor{altoshade}
 \cellcolor{white} & AfroXLMR-L-76L (M) & 0.844 & 0.696 & 0.763 & 0.686 & \textbf{0.837} & 0.754 & \textbf{0.989} & 0.844 & 0.911 & 0.479 & \textbf{0.936} & \textbf{0.634} \\
\midrule
\multirow{6}{*}{fr}
 & OSS-SG-120B & 0.728 & 0.683 & 0.705 & 0.633 & 0.682 & 0.657 & 0.987 & 0.932 & 0.959 & 0.200 & 0.577 & 0.297 \\
 & OSS-SG-20B & 0.751 & 0.648 & 0.695 & 0.626 & 0.733 & 0.675 & \textbf{0.989} & 0.902 & 0.944 & 0.165 & 0.654 & 0.264 \\
 & LFTW & 0.694 & 0.665 & 0.679 & 0.603 & 0.635 & 0.619 & 0.985 & 0.689 & 0.811 & 0.058 & 0.654 & 0.107 \\
 & P-API & 0.726 & 0.536 & 0.617 & 0.564 & 0.748 & 0.643 & 0.986 & 0.470 & 0.636 & 0.041 & \textbf{0.769} & 0.078 \\
\rowcolor{altoshade}
 \cellcolor{white} & CamemBERT-large & 0.809 & \textbf{0.725} & \textbf{0.765} & \textbf{0.697} & 0.787 & \textbf{0.739} & 0.986 & \textbf{0.972} & \textbf{0.979} & \textbf{0.359} & 0.538 & \textbf{0.431} \\
\rowcolor{altoshade}
 \cellcolor{white} & AfroXLMR-L-76L (M) & \textbf{0.832} & 0.496 & 0.621 & 0.582 & \textbf{0.875} & 0.699 & 0.981 & 0.951 & 0.966 & 0.189 & 0.385 & 0.253 \\
\midrule
\multirow{5}{*}{sw}
 & OSS-SG-120B & 0.712 & \textbf{0.924} & \textbf{0.804} & \textbf{0.758} & 0.390 & 0.515 & 0.977 & \textbf{0.912} & \textbf{0.943} & 0.000 & 0.000 & 0.000 \\
 & OSS-SG-20B & 0.701 & 0.822 & 0.757 & 0.595 & 0.427 & 0.498 & 0.975 & 0.875 & 0.922 & 0.000 & 0.000 & 0.000 \\
 & LFTW & 0.701 & 0.761 & 0.730 & 0.546 & 0.469 & 0.504 & 0.976 & 0.891 & 0.931 & 0.000 & 0.000 & 0.000 \\
\rowcolor{altoshade}
 \cellcolor{white} & AfroXLMR-base & 0.835 & 0.683 & 0.751 & 0.601 & 0.780 & \textbf{0.679} & \textbf{0.992} & 0.891 & 0.938 & \textbf{0.118} & \textbf{0.667} & \textbf{0.200} \\
\rowcolor{altoshade}
 \cellcolor{white} & AfroXLMR-L-76L (M) & \textbf{0.849} & 0.627 & 0.721 & 0.573 & \textbf{0.817} & 0.674 & \textbf{0.992} & 0.891 & 0.938 & \textbf{0.118} & \textbf{0.667} & \textbf{0.200} \\
\bottomrule
\end{tabular}
}
\caption{Per-class precision (P), recall (R), and F1 (Non-TFGBV vs.\ TFGBV) by model on the full \texttt{ALTO} dataset ($n{=}2{,}117$) and on \texttt{Bench-test}, for Arabic (ar), French (fr), and Swahili (sw). Zero-shot systems are scored on every ALTO item; \colorbox{altoshade}{shaded} rows are encoder models fine-tuned on ALTO, whose ALTO values are 5-fold CV predictions (inter-fold $\sigma\le 0.08$). AfroXLMR-large-76L was trained as a single multilingual model (M). \textbf{Bold} = best per column and language.}
\label{tab:main_results_appendix}
\end{table*}

\newpage
\subsection{Fine-tuning Significance Test} \label{app:significance_test}
\begin{table}[h]
\centering
\small
\setlength{\tabcolsep}{3pt}
\resizebox{\columnwidth}{!}{%
\begin{tabular}{l rr rr rr}
\toprule
 & \multicolumn{2}{c}{Arabic} & \multicolumn{2}{c}{French} & \multicolumn{2}{c}{Swahili} \\
\cmidrule(lr){2-3}\cmidrule(lr){4-5}\cmidrule(lr){6-7}
System & F1 [95\% CI] & $\Delta$F1 [95\% CI] & F1 [95\% CI] & $\Delta$F1 [95\% CI] & F1 [95\% CI] & $\Delta$F1 [95\% CI] \\
\midrule
OSS-SG-120B & .73 {\scriptsize[.69,.78]} & -.08$^{*}$ {\scriptsize[-.13,-.04]} & .66 {\scriptsize[.62,.69]} & -.08$^{*}$ {\scriptsize[-.12,-.04]} & .52 {\scriptsize[.45,.57]} & -.16$^{*}$ {\scriptsize[-.23,-.10]} \\
OSS-SG-20B & .72 {\scriptsize[.67,.77]} & -.10$^{*}$ {\scriptsize[-.15,-.05]} & .68 {\scriptsize[.64,.71]} & -.06$^{*}$ {\scriptsize[-.10,-.03]} & .50 {\scriptsize[.44,.55]} & -.18$^{*}$ {\scriptsize[-.24,-.12]} \\
LFTW & .69 {\scriptsize[.64,.73]} & -.13$^{*}$ {\scriptsize[-.18,-.08]} & .62 {\scriptsize[.58,.66]} & -.12$^{*}$ {\scriptsize[-.16,-.08]} & .50 {\scriptsize[.45,.56]} & -.17$^{*}$ {\scriptsize[-.23,-.12]} \\
P-API & .72 {\scriptsize[.67,.76]} & -.10$^{*}$ {\scriptsize[-.15,-.05]} & .64 {\scriptsize[.60,.68]} & -.10$^{*}$ {\scriptsize[-.13,-.06]} & -- & -- \\
ALTO fine-tuned (mono)$^{\dagger}$ & .82 {\scriptsize[.78,.85]} & -- & .74 {\scriptsize[.71,.77]} & -- & .68 {\scriptsize[.63,.72]} & -- \\
ALTO fine-tuned (multi) & .75 {\scriptsize[.72,.79]} & -.06$^{*}$ {\scriptsize[-.10,-.03]} & .70 {\scriptsize[.67,.73]} & -.04$^{*}$ {\scriptsize[-.07,-.01]} & .67 {\scriptsize[.63,.72]} & -.01 {\scriptsize[-.04,+.03]} \\
\bottomrule
\end{tabular}}
\caption{TFGBV-class F1 of each system on all \texttt{ALTO-full} items (ar $n$=584, fr $n$=898, sw $n$=635) with 95\% item-bootstrap CIs, and the paired bootstrap difference $\Delta$F1 from the monolingual ALTO fine-tuned encoder on the same items (negative means ALTO fine-tune is better), with its 95\% percentile CI. $^{*}$: the CI of $\Delta$F1 excludes 0 (two-sided bootstrap $p$, Holm-corrected within the language, $p<0.05$). $^{\dagger}$ denotes comparison model. P-API does not support Swahili.}
\label{tab:fn_f1_boot}
\end{table}

\twocolumn
\section{Annotation} \label{app:annotation}

\subsection{Active Learning Details} \label{app:al_details}
The first batch of annotations ($\mathcal{B}_1$) was bootstrapped from $\mathcal{U}$ to sample a diverse set of labels. The second batch ($\mathcal{B}_2$) was sampled using a mix of diversity and uncertainty approaches implemented with the $k$-Nearest Neighbour algorithm, which selects $k$ nearest and farthest neighbours \citep{said_user-centric_2013}. For $\mathcal{B}_3$ and $\mathcal{B}_4$, a pre-trained language model (PLM) was fine-tuned on data from previous batches and instances exhibiting the highest uncertainty or diversity were selected for annotation in $\mathcal{B}_{3+}$.

Normalized entropy $E(x)$ for prediction uncertainty was defined as:
\begin{equation}
    E(x) = \frac{- \sum_{i=1}^{K} p_i(x) \log p_i(x)}{\log K}
\end{equation}
where $p_i$ is the predicted probability of class $i$ for data point $x\in\mathcal{U}$, and $K$ is the total number of classes. As $E(p)$ is normalized by maximum entropy $\log K$, the resulting values are between 0 and 1, corresponding to minimum and maximum entropy. The samples in $\mathcal{U}$ were ranked by $E(x)$ and the top N samples were selected using stratified sampling based on the predicted labels. The batch size and N were determined by the researchers based on the previous round's performance relative to the annotation budget of 1,000 annotations.

Diversity was achieved through activation uncertainty, leveraging the transformer architecture of PLMs to identify diverse candidates for annotation. In a transformer-based PLM, the feature space evolves during fine-tuning, whereas in traditional AL it remains static \citep{wang_deep_2024}. We selected candidate instances at each model uncertainty round ($\mathcal{B}_t$ where $t>2$) by computing the inverse $L2$-norm of the last hidden layer of the PLM and selecting $k$ candidates from $\mathcal{U}$ with the lowest values. This corresponds to the least-confident semantic representation in the current feature space and to potentially out-of-distribution instances \citep{zhang_active_2017}. The $L2$-norm $A(x)$ was calculated as:

\begin{equation}
    A(x) = \frac{1}{\|h(x)\|_2 + \epsilon}
\end{equation}
where $h(x)$ is the activations for the last hidden layer of the PLM when predicting the label of data point $x\in\mathcal{U}$ and $\epsilon$ is a small positive constant to avoid division by zero for very low activations. 
\subsection{Annotator Demographics}
Each language in our dataset was annotated by two annotators and reviewed by an expert annotator. The annotator demographics are presented here.

\para{Levantine Arabic.} Three female annotators between the ages of 25--30 years old were recruited. Two annotators are from Jordan, and one is from Lebanon. Two annotators hold undergraduate degrees, and one holds a master's degree.

\para{Swahili.} Four female annotators between the ages of 25--35 years old were recruited. The additional annotator was due to a change of one annotator after $\mathcal{B}_2$ of annotation. All annotators were from Kenya. They all hold an undergraduate degree. 

\para{African French.} Two female annotators were recruited, one from Benin and one from Nigeria. All hold an undergraduate degree. 

\subsection{Recruitment and Compensation}
All main annotators were members of or consultants engaged by the partner CSOs, and the expert annotators were native speakers and researchers on the study. We provided financial support to the CSOs that included compensation for the annotation work. The partner CSOs were from Jordan, Kenya, Benin and Nigeria. 

\subsection{Annotation Guidelines}
The complete annotation guideline is available on the project repository. 

\section{Experimental and Computing Setup} \label{app:exp_setup}

\subsection{Benchmarking Experiment}
The benchmarking experiment design varied by model architecture type. All experiments were run using the HuggingFace Transformers package \citep{wolf-etal-2020-transformers}. 

\para{Decoder-only Models.} The GPT-OSS-Safeguard models were accessed through the AWS Bedrock platform.\footnote{\url{https://aws.amazon.com/bedrock/}} The inference server location was set to Ireland to ensure processing of all data within a GDPR region. The temperature of the models were set to 0.00.

\para{Encoder-only Models.} Inference by the LFTW model \citep{Vidgen2021} was completed on Google Colab using an A100 GPU. The Perspective API was accessed through Google Cloud.\footnote{\url{https://developers.google.com/codelabs/setup-perspective-api}} Arabic is only supported under the \lstinline{TOXICITY} attribute, which returns a probability between 0 and 1 for the presence of toxic content. The predicted probabilities were binarized using a threshold, with scores above the threshold assigned a positive TFGBV label and scores at or below assigned a negative label. Thresholds between 0.1 and 0.9 were evaluated using F1, with 0.2 chosen as the final label threshold. 

\subsection{Benchmarked Models}\label{app:benchmark_models}
\begin{description}\vspace{-1ex}
  \setlength{\itemsep}{0pt}
  \setlength{\parskip}{0pt}
  \setlength{\parsep}{0pt}
  \setlength{\topsep}{0pt}
    \raggedright
    \item[GPT-OSS-Safeguard (OSS-SG-20B/120B)] Safety fine-tuned decoder-only models from \citet{openai_introducing_2025} 
    \item[Perspective API (P-API)] Google Jigsaw's content moderation model, distilled encoder model\footnote{\url{https://perspectiveapi.com/}}
    \item[Learning-from-the-Worst (LFTW)] A widely used fine-tuned encoder model for harmful content moderation by \citet{Vidgen2021}
\end{description}

\subsection{Fine-tuning Models}\label{app:finetune_models}
ALl fine-tuning models were pre-trained language models (PLM) of the encoder architecture.
\begin{description}\vspace{-1ex}
  \setlength{\itemsep}{0pt} 
  \setlength{\parskip}{0pt}
  \setlength{\parsep}{0pt}
  \setlength{\topsep}{0pt}
    \raggedright
    \item[MARBERTv2 (Arabic)] BERT model fine-tuned on a large Arabic Twitter corpus \citep{AbdulMageed2021} 
    \item[AfroXLMR-base (Swahili)] A multilingual RoBERTa model fine-tuned on 17 African languages including Swahili \citep{alabi-etal-2022-adapting} also benchmarked by \citet{Muhammad2025}
    \item[AfroXLMR-large-76L (multilingual)] A larger model than AfroXLMR-base trained on 76 African languages \citep{adelani-etal-2024-sib}
    \item[CamemBERT (French)] A French fine-tuned BERT model \citep{martin2020camembert}
\end{description}
\subsection{Fine-tuning Experiment}

Each encoder is fine-tuned end to end with a three-head classifier. The pooled or \texttt{[CLS]} output passes through a dropout layer (p=0.1) and a single linear classifier layer with three sigmoid heads for
\emph{gendered content}, \emph{hate} and \emph{TFGBV}. The TFGBV prediction is the third head at a threshold of 0.5, while the other two heads are auxiliary training signals only. The loss is binary cross-entropy with a per-label positive-class weight $n_{\text{neg}}/n_{\text{pos}}$ of the training set, summed over the three heads. The hyperparameters were:
\begin{description}\vspace{-1ex}
  \setlength{\itemsep}{0pt}
  \setlength{\parskip}{0pt}
  \setlength{\parsep}{0pt}
  \setlength{\topsep}{0pt}
    \item Learning rate $= 2\times10^{-5}$ for all, $5\times10^{-6}$ for AfroXLMR-large-76L
    \item Optimizer: AdamW (weight decay 0.01), 10\,\% linear warm-up then linear decay
    \item Epochs $= 10$, last epoch kept (no early stopping or model selection)
    \item Batch size $= 32$
    \item Max token length $= 128$
\end{description}

\subsection{Computing Resources}
All fine-tuning experiments were done on HuggingFace jobs using an A100-small GPU. 5-fold cross-validation of fine-tuning took between 30 minutes for the monolingual models and 2 hours for the multilingual models.

\newpage
\onecolumn
\section{TFGBV Taxonomy}
\textcolor{red}{\textbf{Content warning:} This table contains examples of explicit hate speech, slurs, sexual content, and threats of violence that may be distressful to readers. Please exercise caution.}
\label{sec:taxonomy}
{\footnotesize
\renewcommand{\arraystretch}{1.2}
\setlength{\LTpre}{0pt}
\setlength{\LTpost}{0pt}
 
\begin{longtable}{@{}p{2.6cm} p{2.6cm} p{4.2cm} p{6.0cm}@{}}
\label{tab:taxonomy} \\
\toprule
\textbf{Tier~1} & \textbf{Tier~2} & \textbf{Definition} & \textbf{Examples} \\
\midrule
\endfirsthead
 
\multicolumn{4}{l}{\textit{\tablename~\thetable\ -- continued from previous page}} \\
\toprule
\textbf{Tier~1} & \textbf{Tier~2} & \textbf{Definition} & \textbf{Examples} \\
\midrule
\endhead
 
\midrule
\multicolumn{4}{r}{\textit{continued on next page}} \\
\endfoot
 
\bottomrule
\endlastfoot
 
Non-TFGBV & Non-TFGBV & Content that contains tech-facilitation and gendered content but does not commit violence or harm against the target. This could be content that is neutral, factual, or encouraging. & \textit{She has made us proud Kenyans especially women, balancing career and being a Mum and wife, is not easy. Congratulations to <USER>} \newline \textcolor{gray}{\rule{1cm}{0.4pt}} \newline \textarabic{سبحان الله أغلب التعليقات الظالمة والي تخلو من المراعاة اصحاباها من نفس جنس هالمهندسة الله يقويها ويفرجها عليها} \newline [EN] \textit{Glory be to God! Most of the unfair and insensitive comments are from people of the same gender as this engineer. May God strengthen her and ease her suffering.}  \newline \textcolor{gray}{\rule{1cm}{0.4pt}} \newline \textit{Le respect pour la femme n'est pas une question de vente mais d'éducation. Une fille peut vite coucher avec toi alors que pendant se temps, elle fait poiroter un autre. Cela ne fait pas pour autant d'elle une fille facile ou qui se vend gratuitement.} \newline [EN] \textit{Respect for women isn't a question of transaction, but of upbringing. A girl can sleep with you quickly while, at the same time, keeping someone else waiting. That doesn't make her an 'easy' girl or one who 'gives it away for free.}  

\\
\midrule
Harassment and Hate Speech & Derogation & Explicit misogynistic or abusive content directed towards or about one or a group of targets. This may be expressed as subjective opinions, baseless accusations, negative descriptions, or attacks. & \textit{She is a lying whore and has never said anything true while in office.} 
\newline \textcolor{gray}{\rule{1cm}{0.4pt}} \newline \textit{A hopeless whore who used to be fucked in parliament's toilets, the reason why she remains enamored by politicians\ldots Tusimuone kesho Roysambu wallai. Umbwa nyinyi} \newline [EN] \textit{Let's not see them tomorrow in Roysambu, I swear. You dogs.} 
\newline \textcolor{gray}{\rule{1cm}{0.4pt}} \newline \textit{Personne te respecte car t'aime une salope que t'appelle ta femme...} \newline [EN] \textit{Nobody respects you because you love a slut that you call your wife} \\
 
 & Animosity & Content that may not necessarily be explicit or have a clear target but expresses misogyny or are harmful towards gendered targets. Also known as covert or implicit misogyny. These may be casual use of gendered slurs, stereotyping, or backhanded compliments, condescending messages or disguised as memes. & \textit{Women should be cherished like flowers.} \newline \textcolor{gray}{\rule{1cm}{0.4pt}} \newline \textit{Kwanini miaka ya sasa wanawake wanafanya umalaya bado wakiwa na umri mdogo tofauti na zamani?} \newline [EN] \textit{Why is it that nowadays women engage in prostitution while still very young, unlike in the past?} \newline \textcolor{gray}{\rule{1cm}{0.4pt}} \newline \textarabic{مش حضرتك لبتقرري هيدا شي انتي فيكي تقرري شو تلبسي وكيف تتمكيجي مش اكتر} \newline [EN] \textit{You're not the one who can decide such a thing, you can just decide what to wear and how to do your makeup, nothing more.} 
 \newline \textcolor{gray}{\rule{1cm}{0.4pt}} \newline \textit{le secret avec les femmes, c’est de leur parler uniquement en période d’ovulation, sinon le reste du temps elles ne sont bonnes qu’à péter les couilles} \newline [EN] \textit{The secret with women is to only talk to them during ovulation, the rest of the time they are only good for getting on your nerves}\\

 & Objectification and Dehumanization & Harmful characterisation of gendered targets as non-human entities (e.g.\ insects, animals, disease or refuse), or reduction to sexual objects. Harmful comments about a gender minority's bodies, clothing choices, or sexual lives can also be considered objectifications. & 
\textcolor{gray}{\rule{1cm}{0.4pt}} \newline \textit{Feminists are Western cancer dumped on us.} 
 \newline \textcolor{gray}{\rule{1cm}{0.4pt}} \newline \textit{Tafuta mumama slightly used wa 31 yrs hapo alafu first born ako 3yrs na make sure amekula character development proper. Hajakula pipe kama miezi nane, kula vizuri alafu shika hiyo jadui, usiambie mtu\ldots} \newline [EN] \textit{Find yourself a slightly used woman around 31 years old, whose first-born is about 3 years old. Make sure she's gone through proper character development. She hasn't had sex for around eight months --- sleep with her well, then hold onto that magic, and don't tell anyone.} 
 \newline \textcolor{gray}{\rule{1cm}{0.4pt}} \newline \textit{Depuis quand les femmes doivent elles être applaudies pour s’être renseignées sur comment sucer une bite ou se mettre en position de leuvrette.} \newline [EN] \textit{Since when do women deserve applause for learning how to suck a dick and geting into doggy-style position.}\\

 & Sexual Harassment & Unwanted content of a sexual nature with the purpose or effect of violating the dignity of a person, in particular when creating an intimidating, hostile, degrading, humiliating or offensive environment. & \textit{Anakaa like some babe naeza weka mimba just to slow her down a little\ldots hahaha} \newline [EN] \textit{She looks like the kind of girl I'd get pregnant just to slow her down a little\ldots hahaha} \\
\midrule
Threats and Incitement of Harm and Violence & Threats and Incitement of Physical Violence & Expressing intent, willingness, desire, or the encouragement of others, to commit physical violence. This could include, but is not limited to: killing, beating, harming. & \textarabic{والله والله والله العظيم وين بلاقيك بدي اقتلك قواص ذبح حيالله} \newline [EN] \textit{I swear by God that wherever I find you I will shoot you or slay you.} \newline \textcolor{gray}{\rule{1cm}{0.4pt}} \newline \textit{Mwambie ajidishi. Umbwa!} \newline [EN] \textit{Tell her to kill herself. Dog!} \newline \textcolor{gray}{\rule{1cm}{0.4pt}} \newline \textit{if you can get home addresses for scumfucks, my knife is *awfully* lonely lately.} \newline \textcolor{gray}{\rule{1cm}{0.4pt}} \newline \textit{Il faut souvent frapper ta femme ou ta copine pour qu’elle sache que tu es le chef} \newline [EN] \textit{It is necessary to beat your wife or girlfriend often so she knows you’re the boss.}\\
 & Threats and Incitement of Sexual Violence & Expressing intent, willingness, desire, or the encouragement of others, to commit sexual violence. This could include, but is not limited to: raping, molestation, or other sexual assaults. & \textit{Raping her would put her in her place.} \newline \textcolor{gray}{\rule{1cm}{0.4pt}} \newline \textarabic{و سوف نغتصبك أيضا و معك} J1 \textarabic{الواطية} \newline [EN] \textit{We will rape you and rape degenerate J1 as well.} \\
 & Damning & Wishes to a deity or spiritual power (e.g.\ God) to cause harm, such as illness or death, on the target. & \textarabic{الله يلعنك ويحرقك ويبعتلك سرطان براسك} \newline [EN] \textit{May God curse and burn you and put cancer in your brain.} \\
\midrule
Doxxing & Full or Partial Disclosure & Public release of personally identifiable data or private information of a target. This could include, but is not limited to: legal name, address, phone number, workplace, and private documents or correspondence. & \textit{Her address is XXX road. Go get her!} \\
 & Threat or Incitement of Doxxing & Expressing intent, willingness, desire, or encouraging others to disclose personally identifiable data or private information of a target. This could include, but is not limited to: legal name, address, phone number, workplace, and private documents or correspondence. & \textit{yeah i wish eron would release all the info he's referencing} \\
 
\end{longtable}
}
\clearpage

\includepdf[pages={1}, pagecommand={\section{GPT-OSS-Safeguard Prompt}\label{app:oss_prompt}}]{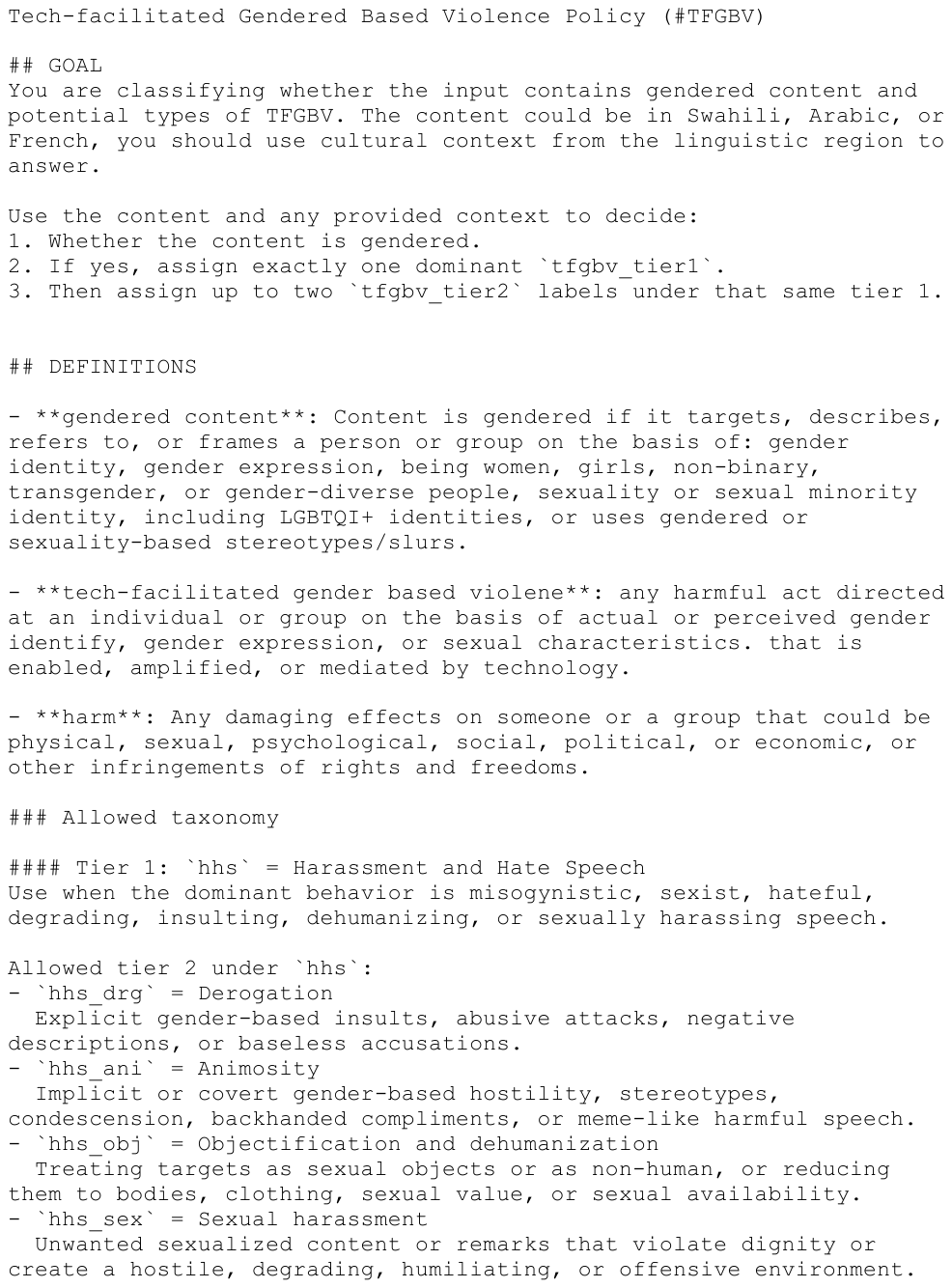}
\includepdf[pages=2-, pagecommand=\thispagestyle{plain}]{src/gpt_oss_prompt.pdf}

\end{document}